\documentclass{article}

\usepackage[final]{corl_2024} %
\usepackage{graphicx}
\usepackage{caption}
\usepackage{subcaption}
\usepackage{pifont} %
\usepackage{threeparttable}
\usepackage{appendix}

\newcommand{\cmark}{\ding{51}} %
\newcommand{\xmark}{\ding{55}} %

\title{Object and Contact Point Tracking in Demonstrations Using 3D Gaussian Splatting}

\author{
  Michael~Büttner \\
  Bielefeld University, Germany \\
  \texttt{mbuettner@techfak.uni-bielefeld.de} \\
  \And
  Jonathan Francis \\
  Bosch Center for AI, USA; \\
  Carnegie Mellon University, USA \\
  \AND
  Helge Rhodin \\
  Bielefeld University, Germany \\
  \And
  Andrew Melnik \\
  Bremen University, Germany \\
}

\begin{document}
\maketitle

\begin{abstract}
This paper introduces a method to enhance Interactive Imitation Learning (IIL) by extracting touch interaction points and tracking object movement from video demonstrations. The approach extends current IIL systems by providing robots with detailed knowledge of both where and how to interact with objects, particularly complex articulated ones like doors and drawers. By leveraging cutting-edge techniques such as 3D Gaussian Splatting and FoundationPose for tracking, this method allows robots to better understand and manipulate objects in dynamic environments. The research lays the foundation for more effective task learning and execution in autonomous robotic systems.

\end{abstract}


\begin{figure}[ht!]
    \centering
    \includegraphics[width=1.0\textwidth]{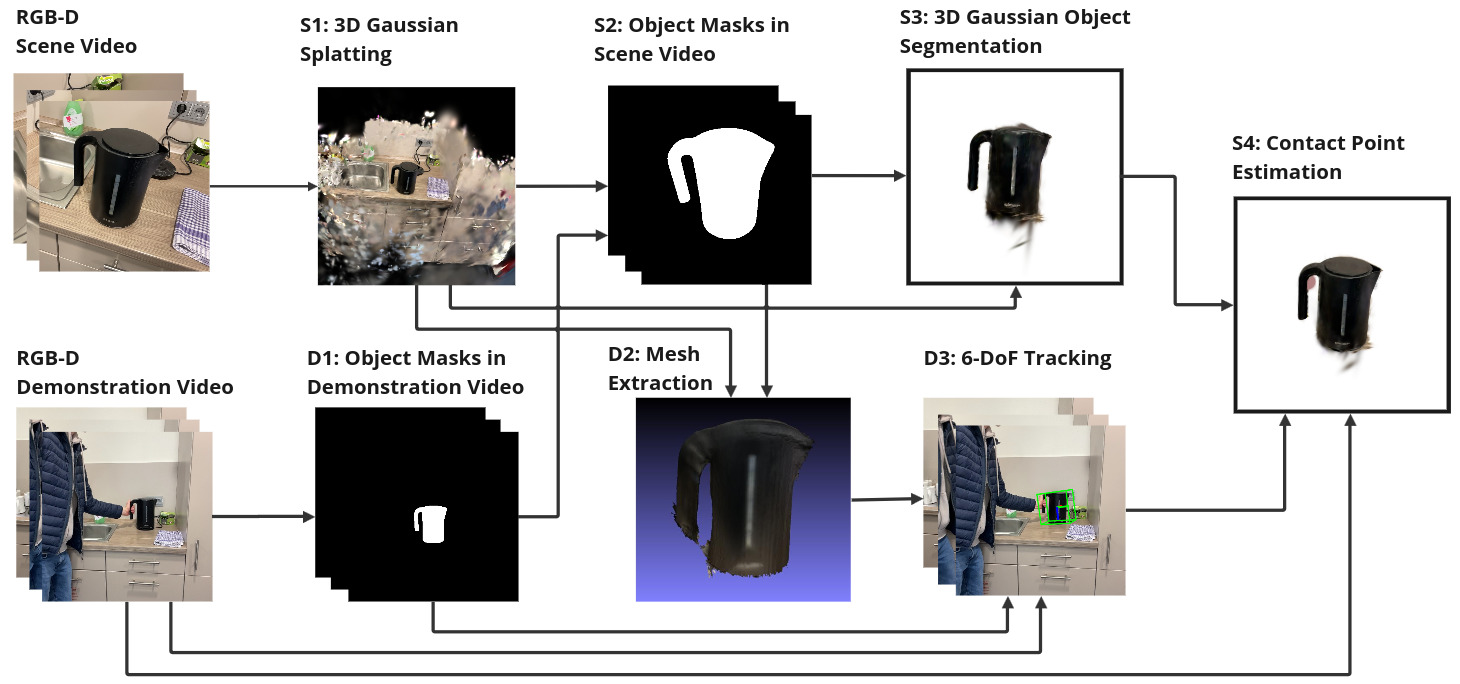}
    \caption{Overview of the pipeline. We start with RGB-D recordings of the scene and the demonstration. We train a 3D Gaussian Splatting \cite{kerbl3Dgaussians} Scene on the scene video, and do object masking on the demonstration video using RAFT \citep{teed2020raft} and SAM 2 \cite{ravi2024sam2}. These masks are used to create object masks of the scene video, which in turn are used to create a mesh using GS2Mesh \cite{wolf2024gs2mesh} and a Gaussian object segmentation using SAGS \cite{hu2024semantic}. The mesh is used to do 6-DoF tracking with FoundationPose \cite{foundationposewen2024}, which in turn is used to estimate contact points. Here, the mesh is visualized by MeshLab \cite{LocalChapterEvents:ItalChap:ItalianChapConf2008:129-136}.}
    \label{fig:overview}
    \vspace{-0.4cm}
\end{figure}

\section{Introduction}
	
Autonomous robotic systems have advanced significantly \cite{yenamandra2024towards,melnik2023uniteam}, but challenges remain in manipulating novel objects and articulated structures like doors or cabinets. Robots must accurately identify, grasp, and move objects within dynamic environments.
Reinforcement Learning (RL) \cite{bach2020learn} and Imitation Learning (IL) \cite{malato2024zero} are widely used approaches, with RL often requiring extensive training and careful reward design, while IL benefits from expert demonstrations but struggles with generalization to unseen scenarios \cite{Sutton1998,Fang2019}. Interactive Imitation Learning (IIL) improves IL by incorporating real-time human feedback \cite{milani2023towards,mikami2024natural}, allowing robots to adjust during execution, but still faces difficulties in tasks requiring nuanced object interaction \cite{celemin2022iil}.

To address this, we propose a method for extracting touch interaction points, or contact points, and tracking object movement from video demonstrations, providing robots with detailed information for manipulating articulated objects. By combining video-based learning with 3D Gaussian Splatting \cite{kerbl3Dgaussians} for 3D scene reconstruction, we create a task-relevant representation of the environment, enabling improved robot performance in complex tasks. This method enhances the IIL framework by offering precise interaction data, laying the groundwork for more autonomous robotic manipulation.

\section{Related Works}
Recent advancements in robotic grasping leverage deep learning models to enable robots to interact robustly with objects in real-world settings. Among foundational approaches, Dex-Net 2.0 introduced a large-scale synthetic grasping dataset to train convolutional neural networks for grasp quality prediction, setting a baseline for efficient grasp point selection through simulated grasping attempts \cite{mahler2017dex}. Building on these concepts, Contact-GraspNet expanded 6-DoF grasp prediction to real-world scenarios using RGB-D data, anchoring 3D points in the point cloud as potential contact points and simplifying the grasp representation into a 4-DoF framework for streamlined inference \cite{sundermeyer2021contact}. Our work builds on these methods by integrating multi-stage estimation for 6-DoF pose tracking and contact prediction, designed to support seamless interaction within a unified, sequential pipeline.

Affordance-based modeling has also provided valuable insights into robotic contact and hand pose estimation. For instance, HRP (Human Affordances for Robotic Pre-Training) annotates "affordance labels" (e.g., 2D contact points, hand poses, and active objects) by analyzing human-object interactions, which enables predictive models for robotic behavior cloning \cite{srirama2024hrp}. Leveraging models like the 100DOH hand-object detector \cite{Shan20} and FrankMocap \cite{rong2021frankmocap}, HRP improves hand-object contact point labeling for robotics, showing the potential of human-inspired affordance prediction for grasp and contact point accuracy. While HRP primarily addresses video-based affordance prediction in 2D images, our approach extends this by integrating contact detection into a 3D reconstruction of the object, enhancing robustness to occlusions and object rotations.

Additionally, Gaussian Splatting has gained traction in robotic applications by enhancing photorealism and creating interactive 3D models in real-time environments. While SplatSim \cite{qureshi2024splatsimzeroshotsim2realtransfer} uses Gaussian Splats in simulators to achieve realistic visual transfer for Sim2Real training, Physically Embodied Gaussian Splatting integrates Gaussian splats with physical priors, dynamically reconstructing 3D environments to enable robotic adaptation to scene changes \cite{abou-chakra2024physically}. In our pipeline, Gaussian Splatting supports updates to 3D object positions, assisting with pose estimation and maintaining model integrity.

\section{Methods}
\label{sec:methods}
\begin{figure}
     \centering
     \begin{subfigure}[b]{0.3\textwidth}
         \centering
         \includegraphics[width=\textwidth]{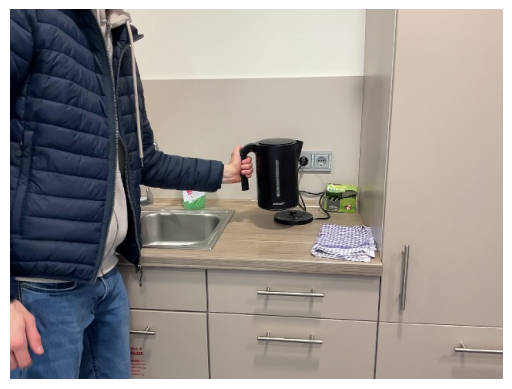}
         \caption{RGB image}
     \end{subfigure}
     \hfill
     \begin{subfigure}[b]{0.3\textwidth}
         \centering
         \includegraphics[width=\textwidth]{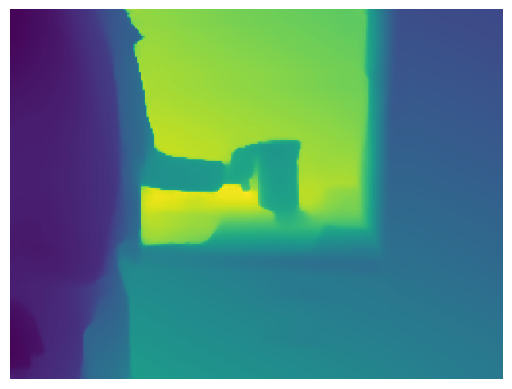}
         \caption{LiDAR depth image}
     \end{subfigure}
     \hfill
     \begin{subfigure}[b]{0.3\textwidth}
         \centering
         \includegraphics[width=\textwidth]{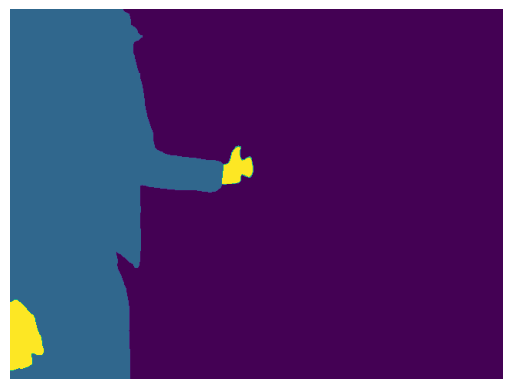}
         \caption{Masks of human and hands}
     \end{subfigure}
     \vfill
     \begin{subfigure}[b]{0.3\textwidth}
         \centering
         \includegraphics[width=\textwidth]{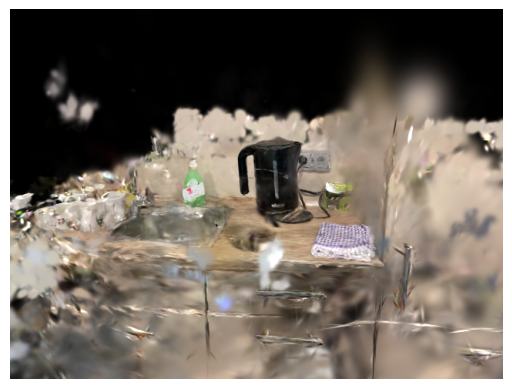}
         \caption{RGB render from the tracking camera position}
     \end{subfigure}
     \hfill
     \begin{subfigure}[b]{0.3\textwidth}
         \centering
         \includegraphics[width=\textwidth]{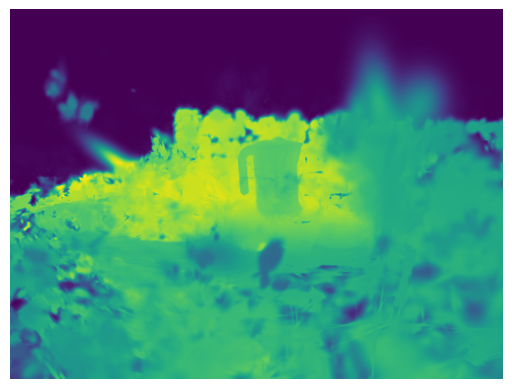}
         \caption{Depth rendering from the tracking camera position}
     \end{subfigure}
     \hfill
     \begin{subfigure}[b]{0.3\textwidth}
         \centering
         \includegraphics[width=\textwidth]{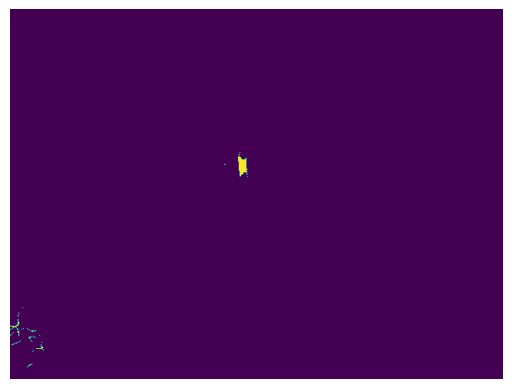}
         \caption{Estimated touch points\newline}
     \end{subfigure}
     \caption{In order to estimate touch points, we calculate the absolute difference between the depth image of the LiDAR camera and the rendered depth image, threshold it, and apply the hand mask that was generated using Grounding DINO \citep{liu2023grounding} + SAM 2 \cite{ravi2024sam2}. This is done for the first 10 frames of contact and then accumulated to find the most probable points.} 
     \label{fig:touch_point_generation}
     \vspace{-0.4cm}
\end{figure}

The basic input consists of two RGB-D videos shot using the Spectacular Rec app \cite{spectacularAI}. The first video is dynamic, capturing the scene from multiple viewpoints with a focus on the object to be manipulated. The second video, called the demonstration video, is a static shot of a human manipulating the object from a fixed camera position. 
These videos allow us to perform 3D Gaussian Splatting \cite{kerbl3Dgaussians} to reconstruct the scene and track the object's 6-DoF pose using FoundationPose \cite{foundationposewen2024}. Contact points are identified through the depth images and object poses. An simple overview can be found in Figure \ref{fig:overview} and a detailed overview can be found in Figure \ref{fig:detailed_overview} in the Appendix. Specifically, the algorithm proceeds as follows:

\begin{enumerate}
	\item In the first step, we use Spectacular AI client to obtain the camera poses and a sparse point cloud from the dynamic scene video. These estimations are then combined with the original 3D Gaussian Splatting training to reconstruct a Gaussian Splatting model of the environment, including the object of interest (Figure \ref{fig:overview} S1). 
	\item In the second step, we export the demonstration video into a format that can be used by FoundationPose. As this tracking method requires object masks for accurate tracking, we first find a bounding box of the manipulated object using RAFT optical flow \citep{teed2020raft}, as shown in Figure \ref{fig:bbox_generation} in the Appendix. We accumulate potential bounding boxes over the video and use a clustering mechanism to find the most probable bounding box. This is used as an input for SAM 2 \cite{ravi2024sam2} to generate these masks (Figure \ref{fig:overview} D1). 
	\item Next, we create masks of the object from the perspective of the scene camera poses (Figure \ref{fig:overview} S2), which can be obtained using the mask from the demonstration video perspective, as well as its camera pose in the scene, which can be obtained using COLMAP \cite{schoenberger2016sfm, schoenberger2016mvs}. The process can be seen in Figure \ref{fig:scene_bbox_generation} in the Appendix.  
    \item With these masks, we create a mesh of the object using GS2Mesh \cite{wolf2024gs2mesh} (Figure \ref{fig:overview} D2) and a segmentation of the Gaussians using SAGS \cite{hu2024semantic} (Figure \ref{fig:overview} S3).  
	\item After obtaining the mesh, FoundationPose is used to track the object's 6DoF pose in camera space (Figure \ref{fig:overview} D3). These poses are translated into the scene space for further use. Using the object's poses and the segmentation obtained by SAGS, the object's trajectory can be visualized in the GS scene. Inaccurate demonstration camera pose estimation can be corrected by shifting the object's estimated starting position to its position in the 3DGS scene.
	\item By utilizing depth images and hand mask data from the demonstration video, along with simple distance thresholding, contact points on the object can be identified, as seen in Figure \ref{fig:touch_point_generation}. We accumulate these points for up to 10 frames of contact and select the most occurring points as the final contact points (Figure \ref{fig:overview} S4).
\end{enumerate}
The processing was carried out on an NVIDIA RTX 3060 Ti GPU of 8GB VRAM. Due to the GPU’s computational constraints, the chosen models balance accuracy and memory efficiency.

\section{Experimental Results}
\label{sec:result}

The video material used in this experiment was recorded utilizing an iPad Pro (11'', 3rd Generation) to capture human-object interaction sequences. The primary focus was to track human touch points on a stationary object during manipulation. The videos were recorded in a controlled environment with consistent lighting, and the camera was positioned at a fixed angle. The video resolution was initially set at 1920x1440 pixels and later reduced to 960x720 pixels to speed up processing without significant loss of detail. Manipulations of 12 different objects were recorded, including articulated objects and portable objects. These videos were recorded to test out the successes and edge cases of this approach. 
The recordings introduced various challenges to assess edge cases: 
\begin{itemize}
\item The jug and mug, which have metallic surfaces, present difficulties due to reflections.
\item The jug, mug and the pincher, also have smooth, textureless surfaces, which adds complexity.
\item The detergent bottle includes a transparent section, making its demonstration video more challenging, especially with significant hand occlusion.
\item The detergent bottle demonstration also includes a large occlusion, as the demonstrator's hand covers a large section of the bottle.
\item The shoe and the brush demonstrations include heavy rotations.
\item The shoe demonstration also shows the shoe's underside, which is not seen in the scene.
\item The cookies box presented a scenario with large displacement, as it was moved far away from the camera during the demonstration.
\item The fridge demonstration video includes a person walking in front of the fridge to test partial occlusion. 
\item The fridge is not fully visible, as the top end of the fridge is outside the camera's view.
\item The cabinet demonstration video has the objects on the counter rearranged compared to the scene video.
\end{itemize}
Each component of the pipeline was evaluated separately to assess individual performance and isolate failure points. Since models operate sequentially, inaccuracies in earlier stages could potentially propagate through the pipeline and affect subsequent outcomes. To mitigate this, we performed manual interventions when possible, enabling isolated evaluation of each step. For 6-DoF pose estimation using FoundationPose, we assessed two approaches: (1) standard pose tracking, which relies on the previous frame’s estimated pose for current pose predictions, and (2) per-frame pose estimation, which independently estimates poses in each frame using object masks. Although standard tracking achieves near-real-time performance, per-frame estimation incurs an 11x slowdown, trading off speed for accuracy. Given the lack of ground-truth object poses, contact point estimations were derived from estimated poses. We calculated the cumulative success by combining the steps from the previous approaches.

\begin{figure}
     \centering
     \begin{subfigure}[b]{0.45\textwidth}
         \centering
         \includegraphics[width=\textwidth]{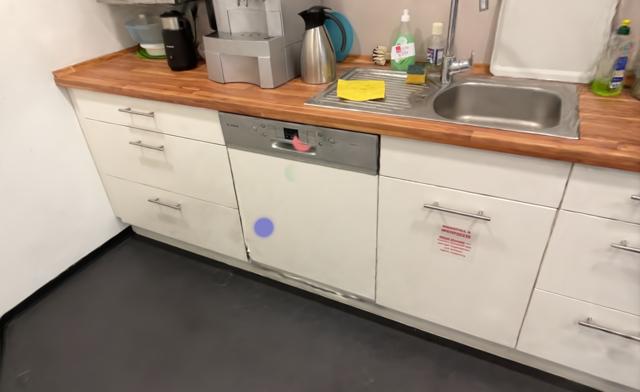}
         \caption{Closed Dishwasher}
         \label{fig:dish1}
     \end{subfigure}
     \begin{subfigure}[b]{0.45\textwidth}
         \centering
         \includegraphics[width=\textwidth]{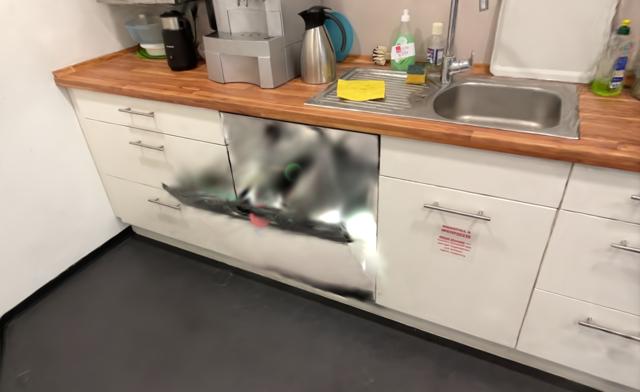}
         \caption{Open Dishwasher}
         \label{fig:dish2}
     \end{subfigure}
     \vfill
	 \begin{subfigure}[b]{0.45\textwidth}
         \centering
         \includegraphics[width=\textwidth]{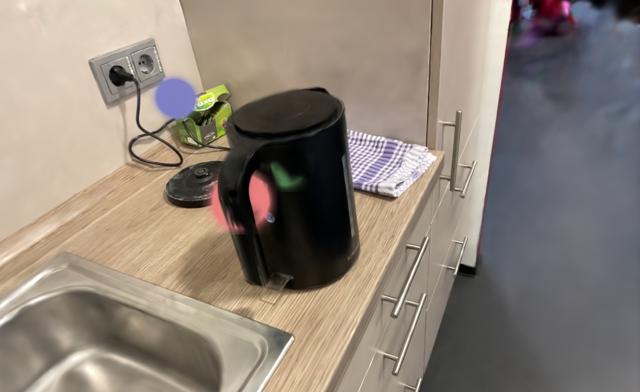}
         \caption{Kettle at the start position}
         \label{fig:kettle1}
     \end{subfigure}
     \begin{subfigure}[b]{0.45\textwidth}
         \centering
         \includegraphics[width=\textwidth]{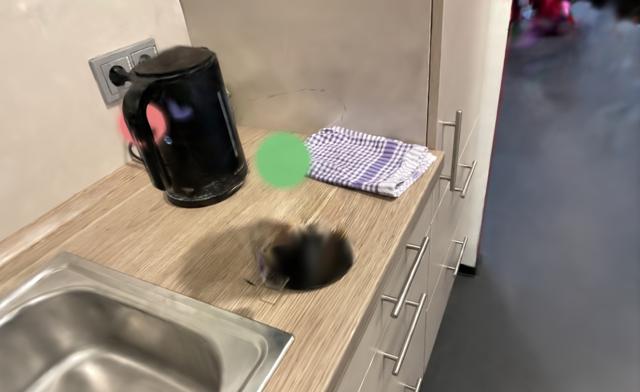}
         \caption{Kettle at the end position}
         \label{fig:kettle2}
     \end{subfigure}
       \vfill
     \begin{subfigure}[b]{0.45\textwidth}
         \centering
         \includegraphics[width=\textwidth]{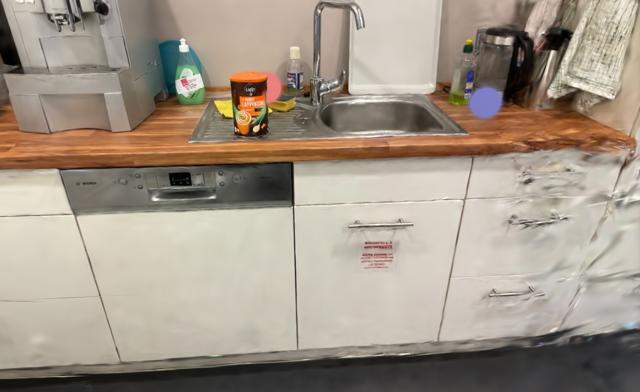}
         \caption{Cappuccino canister at the starting position}
         \label{fig:cappuccino1}
     \end{subfigure}
     \begin{subfigure}[b]{0.45\textwidth}
         \centering
         \includegraphics[width=\textwidth]{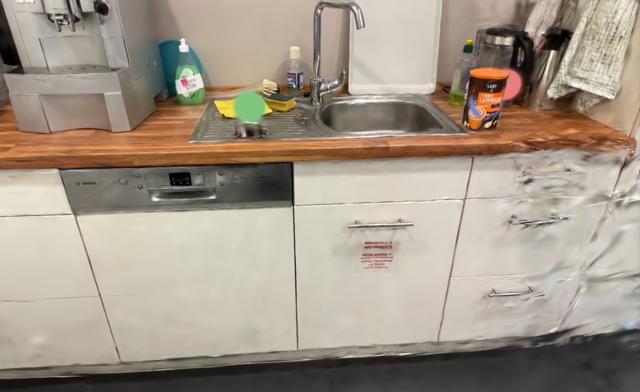}
         \caption{Cappuccino canister at the ending position}
         \label{fig:cappuccino2}
     \end{subfigure}
     \caption{Visuals for successful tracking and contact point estimation episodes. The red spheres stand for the identified contact points, the green spheres for the starting positions, and the blue spheres stand for the end position.} 
     \label{fig:success}
     \vspace{-0.4cm}
\end{figure}

The method produced the most reliable results when applied to large objects with visually distinct features on all sides, such as kettles and cappuccino canisters, as this aids in more consistent pose estimation across frames. Similarly, articulated objects with large flat surfaces, such as fridges or dishwashers, allowed for smoother tracking. These characteristics helped minimize ambiguity in object recognition, leading to more accurate results. Contact point estimation yielded equivalent results on both pose tracking and per-frame pose estimation. Even among episodes where tracking was lost shortly after the manipulation starts, correct contact points were estimated. Visual examples can be found in Figure \ref{fig:success}.

However, since multiple models are applied sequentially, errors or imprecisions in one stage can propagate through the pipeline and negatively impact subsequent steps. While 3D Gaussian Splatting combined with Spectacular AI's pose estimation is generally robust, light reflections on surfaces, such as on the jug and the mug, can introduce artifacts in the texture of the mesh, which in turn affect the accuracy of the 6DoF pose estimation.
When applying RAFT to the demonstration video, the model had difficulty detecting motion on plain surfaces and frequently produced false positives when shadows were present, although outliers in single frames could be filtered out by the bounding box clustering approach. This approaches struggled with doors such as the dishwasher door, as there is less movement on the section near the hinges, leading to smaller bounding boxes. Even when bounding boxes were correctly defined, SAM 2 occasionally failed to consistently generate object masks, with the segmentation breaking down at certain time steps. In the fridge scene, SAM 2 would segment the fridge door's surface while leaving out the handle, which is a critical omission later in the pipeline when contact points are estimated. Even manual input annotation would not lead to successful cases, where both the door and the handle are segmented.
GS2Mesh, which uses stereo vision for depth prediction, sometimes produced faulty depth maps on textureless or reflective surfaces, leading to unnatural deformations or dents in the mesh that could disrupt the 6DoF tracking. 

Furthermore, SAGS's voting system for determining which Gaussians belong to the object requires the entire object to be within the camera's field of view at each frame; otherwise, parts of the object may be excluded, which happened with the upper section of the fridge door. Conversely, relaxing this restriction to give Gaussians outside the camera's frustum neutral votes results in the inclusion of irrelevant Gaussians far from the object.

FoundationPose struggled with objects that lacked distinct features on all sides, such as a jug with a handle. In such cases, it often misestimated the object's orientation, rotating the handle out of view, which creates problems for touch point prediction. Using standard pose estimation, tracking was lost for smaller objects such as the mug, the pitcher and the brush. While all other objects have accurate estimations before manipulation, the brush's estimated position deviated immediately from its true position. The resulting estimations appeared to float in midair. Similarly, occlusions was an issue for standard pose tracking, as the fridge's estimated pose shifted to the right during partial occlusion and the detergent's pose was lost entirely during manipulation. Using per-frame pose estimation improved the 3D position estimation of these objects, but rotation estimation accuracy declined, as some frames received erratic shifts in their rotation. 
Large distances to the camera also become a problem in the cookies box scene. While pose estimation is successful for the first half of the video, it declined in the second half as standard pose tracking lost track of the object, while per-frame pose estimation results jumped far into the background. In the shoe scene, tracking was accurate before the underside of the shoe was shown. Afterwards, the 3D position estimation remained close to the object while rotation accuracy varied.
For transferring the pose estimation into the Gaussian Splatting scene, COLMAP was used. While not shown in this table, COLMAP failed in environments with insufficient keypoints or when the demonstration camera deviated significantly from the scene camera's trajectory.

Lastly, while pose estimation was successful for most portable objects, articulated objects received mixed results. While the dishwasher's contact point was correctly placed on its recessed handle, the other articulated objects' contact points were placed on the surface near the handle.

When combining these steps, we can evaluate the overall success of the approach. Considering a "High-Accuracy Success", where both 3D position and rotation accuracy are highly accurate, two scenes are successful, the kettle scene and the cappuccino scene, resulting in a success rate of 17\%. Here, 4 scenes fail as early as the masking stages, 5 scenes at the pose tracking stage and 1 at the contact point estimation stage. When considering a "Low-Accuracy Success", where rotation accuracy isn't considered, the jug, mug and detergent scenes are included, raising the success rate to 42\%.
\begin{table}[ht]
\centering
\caption{Success for individual components of the pipeline, as well as a combination of all components. "High-Accuracy Success" indicates precise 6DoF pose tracking, while "Low-Accuracy Success" accepts moderate positional alignment without regard for rotational accuracy.}
\label{tab:overall_success}
\scalebox{0.6}{ 
\begin{tabular}{|l||p{1.cm}|p{1.cm}|p{1.cm}|p{1.cm}|p{1.cm}|p{1.cm}|p{1.cm}|p{1.cm}|p{1.cm}|p{1.cm}|p{1.cm}|p{1.cm}||p{2.cm}|}
    \hline
    \hline
    \textbf{}                   & \textbf{Fridge} & \textbf{Cabinet} & \textbf{Dish\-washer} & \textbf{Kettle} & \textbf{Cappu\-ccino} & \textbf{Pitcher} & \textbf{Cookies Box} & \textbf{Shoe} & \textbf{Jug} & \textbf{Mug} & \textbf{Deter\-gent} & \textbf{Brush} & \textbf{Success Rate} \\ \hline \hline
    \multicolumn{14}{|c|}{\textbf{Step-Specific Success}} \\ \hline
    \textbf{Dem. Obj. Masking}  & \xmark*  & \cmark  & \xmark  & \cmark  & \cmark  & \cmark  & \cmark  & \cmark & \cmark  & \cmark  & \cmark  & \cmark   & \textbf{10/12} (83\%)    \\ \hline \hline
    \textbf{Scene Obj. Masking}  & (\xmark)  & \cmark  & \cmark  & \cmark  & \cmark  & \xmark**   & \cmark  & \xmark   & \cmark & \cmark & \cmark & \cmark & \textbf{\phantom{0}9/12} (75\%)    \\ \hline \hline
    \textbf{Pose Tracking} & & & & & & & & & & & & &\\
    \textbf{  Position}      & \cmark-  & \cmark  & \cmark  & \cmark  & \cmark  & \cmark\xmark   & \cmark\xmark  & \cmark-   & \cmark & \xmark & \xmark & \xmark & \textbf{\phantom{0}7/12} (58\%)     \\ 
    \textbf{  Rotation}   & \cmark  & \cmark  & \cmark  & \cmark  & \cmark  & \cmark\xmark   & \cmark\xmark  & \cmark\xmark & \xmark & \xmark & \xmark & \xmark &  \textbf{\phantom{0}5/12} (42\%)   \\ \hline 
\textbf{Per-Frame Pose Est.} & & & & & & & & & & & & &\\
    \textbf{  Position}      & \cmark  & \cmark  & \cmark  & \cmark  & \cmark  & \cmark   & \cmark\xmark  & \cmark   & \cmark & \cmark & \cmark & \cmark & \textbf{11/12} (92\%)    \\ 
    \textbf{  Rotation}   & \cmark\xmark  & \cmark  & \cmark  & \cmark-  & \cmark  & \cmark\xmark   & \cmark\xmark  & \cmark\xmark & \xmark & \xmark & \xmark & \cmark\xmark & \textbf{\phantom{0}4/12} (33\%)    \\ \hline \hline
    \textbf{Contact Point Est.}      & \xmark  & \xmark  & \cmark & \cmark & \cmark  & \cmark   & \cmark  & \cmark   & \cmark & \cmark & \cmark & \xmark & \textbf{\phantom{0}9/12} (75\%)    \\ \hline \hline
    \multicolumn{14}{|c|}{\textbf{Overall Success}} \\ \hline
    \textbf{High-Acc. Success w/} & & & & & & & & & & & & &\\
    \textbf{ Pose Tracking}  & \xmark  & \xmark  & \xmark & \cmark & \cmark  & \xmark   & \xmark  & \xmark   & \xmark & \xmark & \xmark & \xmark & \textbf{\phantom{0}2/12} (17\%)    \\ \hline
    \textbf{Low-Acc. Success w/} & & & & & & & & & & & & &\\
    \textbf{ Per-Frame Pose Est.} & \xmark  & \xmark  & \xmark & \cmark & \cmark  & \xmark   & \xmark  & \xmark   & \cmark & \cmark & \cmark & \xmark & \textbf{\phantom{0}5/12} (42\%)    \\ \hline
\end{tabular}
}
\begin{tablenotes}
\item \cmark\phantom{**} success
\item \cmark-\phantom{* } success, but with less accuracy
\item \cmark\xmark\phantom{*} mixed results, with both accurate and highly inaccurate estimations
\item \xmark\phantom{** } failure
\item (\xmark)\phantom{*} masks capture the object while leaving out the handle
\item \xmark*\phantom{* } segmentation mask appears incorrect for the first few frames, but remains correct for the frames after
\item \xmark**\phantom{ } no masks were generated for earlier frames of the demonstration, but only for later frames
\end{tablenotes}
\end{table}

\section{Discussion}
\label{sec:discussion}
This method offers an innovative approach for extracting contact points and object trajectories from video demonstrations, enabling robots to manipulate articulated and portable objects like dishwashers and kettles.

However, there are technical and practical limitations to consider when applying this approach. FoundationPose yields state of the art results in 6DoF pose estimation tracking but relies heavily on a high-quality mesh reconstructions, which are often challenging to generate. Factors such as object size, texture, shape, distance from the camera, light reflections, and computational power influence the potential quality of the mesh; all of these factors are prominent in real-world environments. The human user would need to ensure minimal occlusion of the object during demonstration, while ideally moving the object slowly and only showing parts of the object viewable in the scene video. This becomes a problem for articulated objects like doors, that move their outer surface outside the camera's view while being opened. Given enough computing power, future work could employ BundleSDF \cite{bundlesdfwen2023} to simultaneously track and reconstruct the mesh. The effectiveness of the approach also depends on specific object characteristics, with the approach performing best on rigid, textured objects while presenting challenges for small, textureless or deformable deformable objects.

Bounding box estimation presents another challenge. While manual assignment is more effective,
automatic detection is the preferred solution. Existing tools like Grounding DINO \cite{liu2023grounding} and the ”100 days of hands” hand-object detector \cite{Shan20} struggle with distinguishing between moving and static parts of articulated objects, often encompassing stationary sections in the bounding box. This limitation significantly impacts grasp detection in complex objects like doors or appliances. For that reason, we chose an optical flow approach to determine the bounding box, as it focuses only on the moving parts of objects. However, this approach faces other challenges with articulated objects, as reducing the motion threshold to account for movement near hinges increases its susceptibility to noise. In our tests it failed on the one articulated object that generated correct contact points. 

Additionally, the system currently requires several minutes per video for processing steps such as masking, mesh extraction, and pose tracking, especially if per-frame pose estimation is used. This processing time makes real-time application challenging, which may limit its immediate feasibility in scenarios requiring rapid response. However, if meshes are pre-computed and standard pose tracking is employed, near real-time application is possible. Otherwise, improvements in algorithmic efficiency or access to more powerful hardware would be necessary for real-time application on mobile robots.

Environment and lighting conditions also pose limitations, as changes in lighting, reflective surfaces, and cluttered scenes can interfere with depth estimation quality and consequently mesh extraction, impacting downstream processes such as pose tracking and object manipulation. However, as long as an accurate demonstration camera pose is maintained, variations in the placement of background objects between the real-world scene and the 3DGS environment, and potentially different lighting conditions, do not affect tracking performance.

Since multiple models are applied sequentially, the pipeline is subject to error propagation, where inaccuracies in early stages, such as camera pose estimation or object masking, can compound in later stages. These accumulated errors can lead to significant deviations in pose tracking and contact point prediction, limiting the overall reliability of the approach in complex scenarios.

Our system identifies contact points by combining hand masks and depth images. This approach works well with portable objects, as the hand generally avoids occluding other surfaces of the object. However, with larger objects like doors, the hand must cover a substantial portion of the object’s surface to reach handles, leading to occasional misalignment between depth data and masks. This misalignment can create inaccurate contact points, which happened in the cabinet scene. Additionally, our contact point predictions are limited to the visible sections of objects, posing a challenge for robotic applications that require precise, encompassing grasps. For example, when attempting to grasp a cylindrical object like a bottle, predicted contact points may only cover one side, often occluded by the hand, whereas successful grasps require more complete object coverage. Future work could address these challenges by incorporating hand pose estimation to infer occluded areas and establish contact points on parts of the object currently out of view. 
Additionally, successful application of these contact points assumes precise end-effector calibration on the robot; any misalignment could further reduce grasping accuracy.

\section{Applicability}
To apply this approach on a mobile robot, a compatible RGB-D camera with IMU recording and a gripper arm would be essential for consistent tracking and object manipulation. 
The robot could first explore its environment to create a 3D Gaussian Splatting scene, which would serve as a digital twin environment. This mapping would allow the robot to not only locate objects, but also orient itself within the environment. This self-localization could potentially bypass the need for COLMAP for demonstration camera pose estimation, enhancing both efficiency and accuracy when tracking its own movement relative to the objects. Additionally, a photorealistic 3DGS reconstruction of the environment enables the robot to infer computer vision models on parts of the environment outside of its current location.

When tasked with actions like opening a door, the robot could analyze human demonstration data to determine where to grasp and to trace the object's trajectory for accurate replication. With accurate 6-DoF pose estimation, the robot could also interpret an object's intended orientation (e.g., ensuring a mug remains upright during transport) and refine its manipulation skills, such as rotating a pitcher to pour liquid into a mug. Moreover, it could track an object's location across sessions, which would allow it to retrieve items or monitor changes in a dynamically rearranged environment - critical for mobile robotics in dynamic, real-world settings.

\section{Conclusion}
\label{sec:conclusion}

This paper presents a novel approach utilizing Gaussian Splatting to infer grasp points and object trajectories from video demonstrations, providing a robust framework for robots to interact with both articulated and portable objects in real-world environments. By reconstructing dynamic scenes and identifying critical touchpoints, this method could enable robots to manipulate everyday items, such as doors and appliances, offering valuable applications in home automation and service robotics.

Despite its promise, the method faces challenges, including reliance on high-quality mesh reconstructions, accurate depth estimations, and bounding box inaccuracies. Addressing these limitations - particularly in automatic object segmentation and mesh generation techniques, while keeping a low computational overhead - is crucial for broader practical applications.

Nevertheless, the potential of this approach is clear. As advancements in scene reconstruction, point tracking, and pose estimation continue to develop, this method could significantly improve robotic autonomy and efficiency. Future work should focus on refining these components and exploring new algorithms that minimize computational complexity, allowing for more effective real-time interaction. Ultimately, this approach opens new avenues for research in robotic manipulation and interaction, pushing the boundaries of what is achievable in home robotics today.

\clearpage
\acknowledgments{The research reported in this paper has been partially supported by DAAD PPP-USA program under project number 57651674, and the German Research Foundation DFG, as part of Collaborative Research Center 1320 Project-ID 329551904 “EASE - Everyday Activity Science and Engineering”, University of Bremen (http://www.ease-crc.org).}

\bibliography{example}  %

\appendix
\renewcommand{\thefigure}{A.\arabic{figure}}
\setcounter{figure}{0}
\section{Appendix}\label{appendix}

\begin{figure}[ht!]
    \centering
    \includegraphics[width=1.0\textwidth]{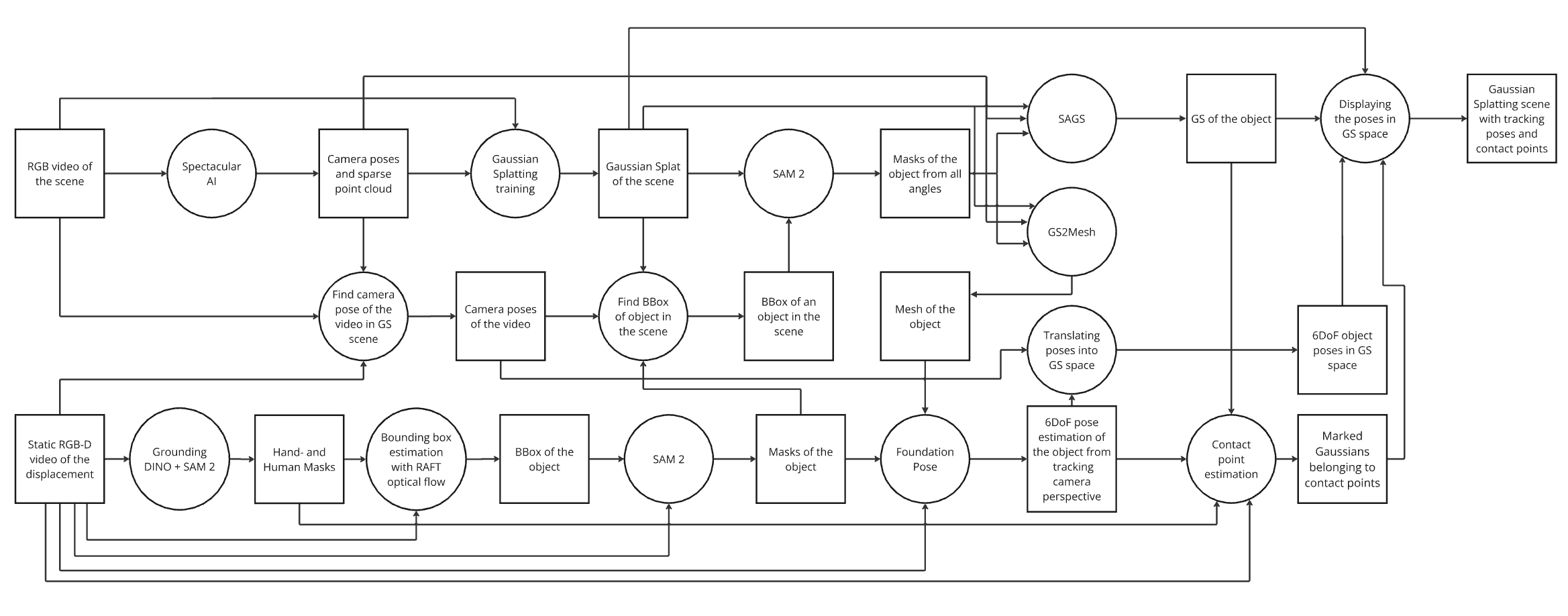}
    \caption{Detailed overview of the process. Rectangles indicate data while circles indicate processes.}
    \label{fig:detailed_overview}
\end{figure}

\begin{figure}[ht]
     \begin{subfigure}[b]{0.3\textwidth}
         \centering
         \includegraphics[width=\textwidth]{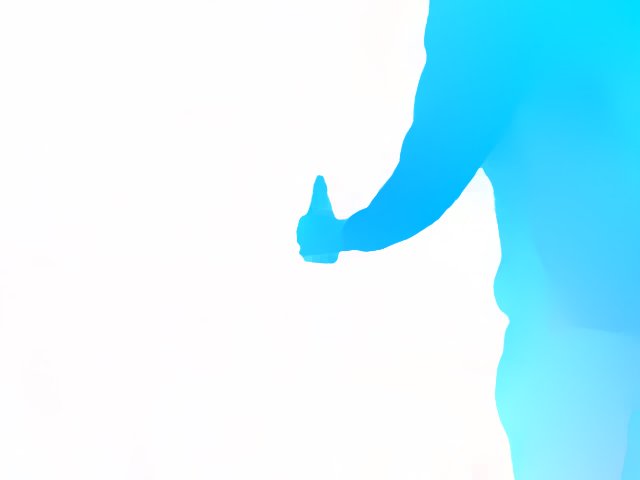}
         \caption{RAFT optical flow output\newline}
     \end{subfigure}
     \hfill
     \begin{subfigure}[b]{0.3\textwidth}
         \centering
         \includegraphics[width=\textwidth]{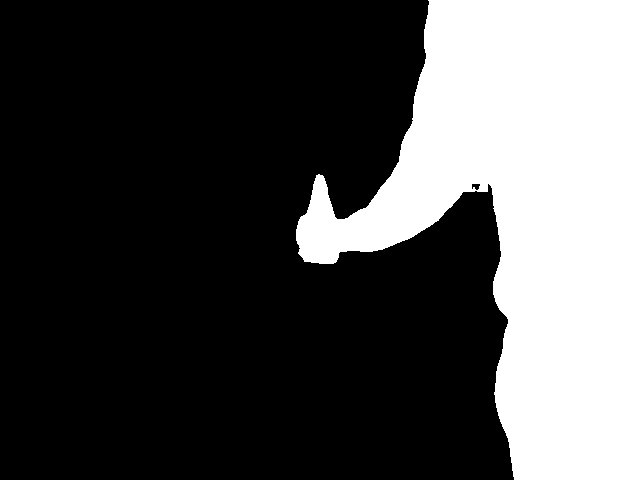}
         \caption{Movement mask based on threshold}
     \end{subfigure}
     \hfill
     \begin{subfigure}[b]{0.3\textwidth}
         \centering
         \includegraphics[width=\textwidth]{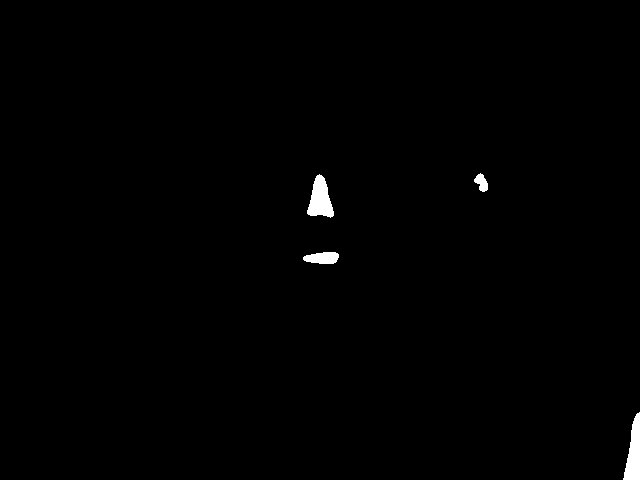}
         \caption{Removing the human mask + opening}
     \end{subfigure}
	 \vfill
     \begin{subfigure}[b]{0.3\textwidth}
         \centering
         \includegraphics[width=\textwidth]{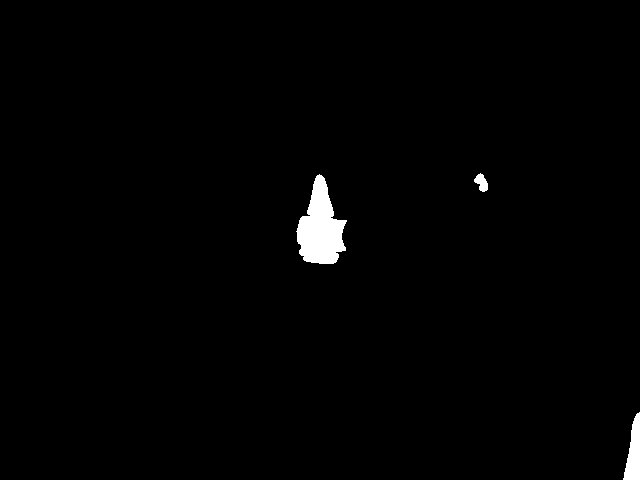}
         \caption{Adding the hand mask\newline}
     \end{subfigure}
     \hfill
     \begin{subfigure}[b]{0.3\textwidth}
         \centering
         \includegraphics[width=\textwidth]{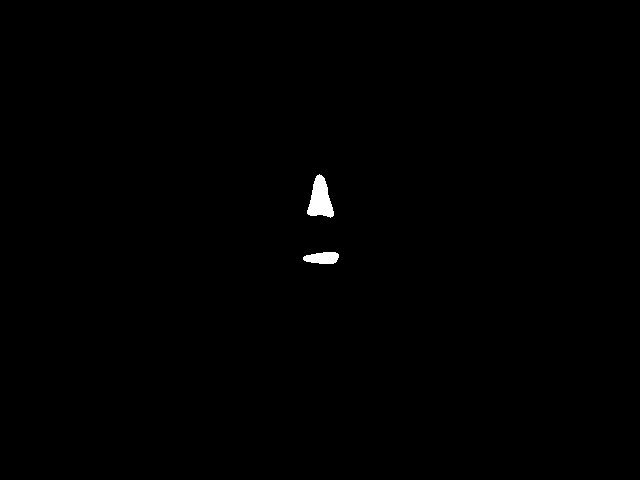}
         \caption{The final determined cluster for this timestep}
     \end{subfigure}
     \hfill
     \begin{subfigure}[b]{0.3\textwidth}
         \centering
         \includegraphics[width=\textwidth]{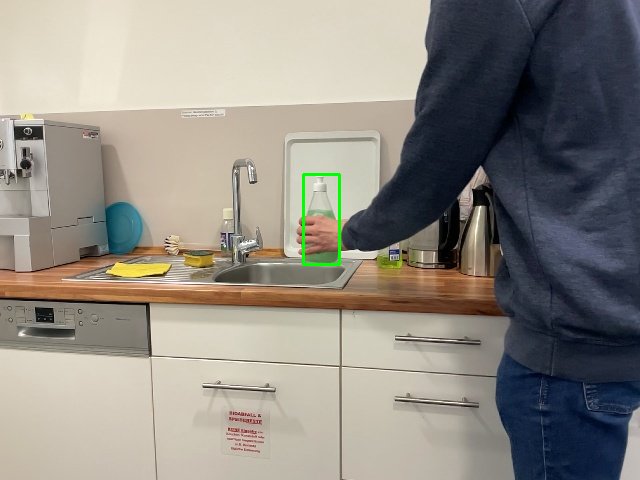}
         \caption{Frame with generated bounding box}
     \end{subfigure}
     \caption{In order to find the bounding box of the moving object, we use RAFT optical flow on all the frames of the demonstration video with a stride of 6, remove the human from the optical flow mask, cluster the bounding boxes by size and choose the frame whose bounding box is closest to the center of the biggest cluster. The human is filtered out using a mask generated by Grounding DINO \citep{liu2023grounding} and SAM 2 \citep{ravi2024sam2}.} 
     \label{fig:bbox_generation}
\end{figure}

\begin{figure}
     \centering
     \begin{subfigure}[b]{0.47\textwidth}
         \centering
         \includegraphics[width=\textwidth]{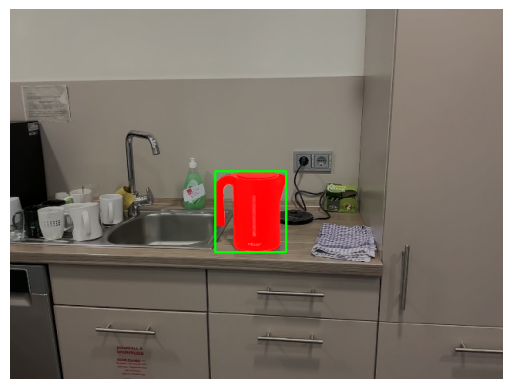}
         \caption{First frame of the demonstration video, with mask applied to it}
     \end{subfigure}
     \hfill
     \begin{subfigure}[b]{0.47\textwidth}
         \centering
         \includegraphics[width=\textwidth]{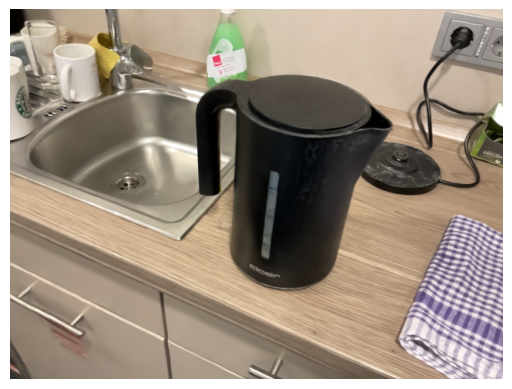}
         \caption{Kettle rendered from the closest scene camera view}
     \end{subfigure}
     \caption{In order to find masks of the object in the scene, we take the first frame of the demonstration video and insert it into the scene video trace at the closest camera position. The bounding box from the demonstration video's object mask is taken as an input. To ensure closer alignment to the underlying representation, the scene frames are rendered from the 3DGS scene instead of taken from the video. } 
     \label{fig:scene_bbox_generation}
\end{figure}

\end{document}